%% file: BEV-LaneDet.tex
\documentclass[10pt,twocolumn,letterpaper]{article}

\usepackage[pagenumbers]{cvpr} 

\usepackage{graphicx}
\usepackage{amsmath}
\usepackage{amssymb}
\usepackage{booktabs}
\usepackage{algorithm}
\usepackage{algorithmic}
\usepackage{multirow}
\usepackage[table,xcdraw]{xcolor}

\usepackage[pagebackref,breaklinks,colorlinks]{hyperref}

\usepackage[capitalize]{cleveref}
\crefname{section}{Sec.}{Secs.}
\Crefname{section}{Section}{Sections}
\Crefname{table}{Table}{Tables}
\crefname{table}{Tab.}{Tabs.}

\def\cvprPaperID{4616} 
\def\confName{CVPR}
\def\confYear{2023}

\begin{document}

\title{BEV-LaneDet: a Simple and Effective 3D Lane Detection Baseline}

\author{Ruihao Wang$^{1,2}$, Jian Qin\thanks{Corresponding author} $^{1}$, Kaiying Li$^{1}$, Yaochen Li$^{2}$, Dong Cao$^{1}$, Jintao Xu$^{1}$\\
$^{1}$HAOMO.AI Technology Co., Ltd.\\
$^{2}$Xi’an Jiaotong University\\
{\tt\small  \{wangruihao, qinjian, likaiying, caodong, xujintao\}@haomo.ai, yaochenli@mail.xjtu.edu.cn} 
}
\maketitle

\begin{abstract}
   3D lane detection which plays a crucial role in vehicle routing, has recently been a rapidly developing topic in autonomous driving. Previous works struggle with practicality due to their complicated spatial transformations and inflexible representations of 3D lanes. Faced with the issues, our work proposes an efficient and robust monocular 3D lane detection called BEV-LaneDet with three main contributions. First, we introduce the \textit{Virtual Camera} that unifies the in/extrinsic parameters of cameras mounted on different vehicles to guarantee the consistency of the spatial relationship among cameras. It can effectively promote the learning procedure due to the unified visual space. We secondly propose a simple but efficient 3D lane representation called \textit{Key-Points Representation}. This module is more suitable to represent the complicated and diverse 3D lane structures. At last, we present a light-weight and chip-friendly spatial transformation module named \textit{Spatial Transformation Pyramid} to transform multiscale front-view features into BEV features. Experimental results demonstrate that our work outperforms the state-of-the-art approaches in terms of F-Score, being 10.6\% higher on the OpenLane dataset and 5.9\% higher on the Apollo 3D synthetic dataset, with a speed of 185 FPS. The source code will released at \url{https://github.com/gigo-team/bev_lane_det}.
\end{abstract}

\section{Introduction}
\label{sec:introduction}
\input{chapter/introduction}


\section{Related Work} 
\label{sec:related}

\input{chapter/related_work}

\section{Methodology} 
\label{sec:methodology}
\begin{figure*}[t]
    \centering
      \includegraphics[width=.9\textwidth]{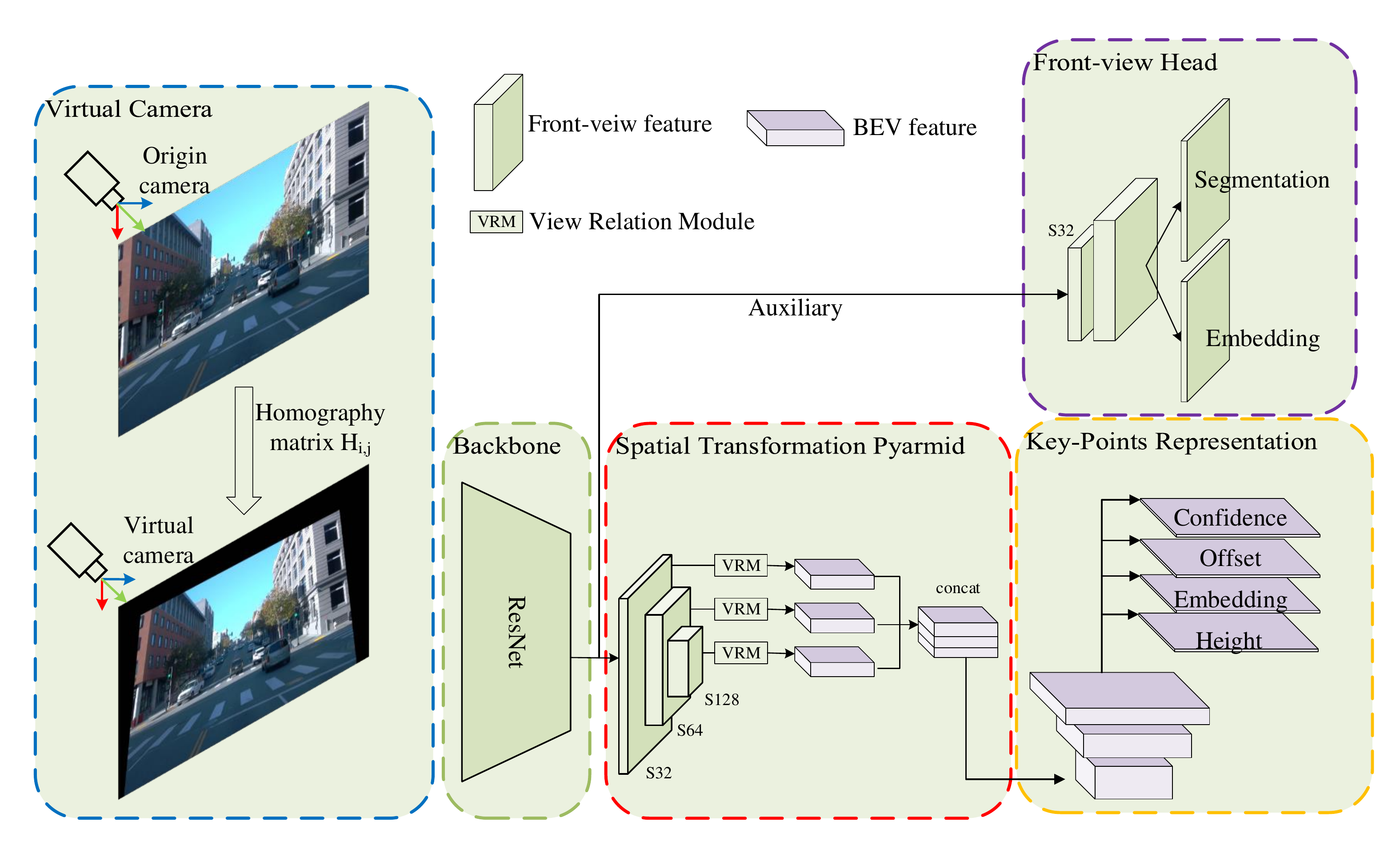}
      \caption{Our network structure consists of five parts: \textit{Virtual Camera}, Backbone, \textit{Spatial Transformation Pyramid}, \textit{Key-Points Representation}, Front-view Head. $S32$: 32x downsampling of the input image.}
    \label{fig:backbone}
\end{figure*}

As shown in Figure \ref{fig:backbone}, the whole network architecture consists of five parts: \textbf{1)} \textit{Virtual Camera}: a preprocessing method for unifying camera intrinsic and extrinsic parameters; \textbf{2)} Front-view Backbone: a front-view features extractor; \textbf{3)} \textit{Spatial Transformation Pyramid}: Projecting  front-view features to BEV features; \textbf{4)} \textit{Key-Points Representation}: a 3D head detector based on key-points; \textbf{5)} Front-view Head: a 2D lane detection head to provide auxiliary supervision.

Firstly, all in/extrinsic parameters of input images are transformed into unified in/extrinsic parameters through the \textit{Virtual Camera}. This process ensures the consistency of the spatial relationship of front-facing cameras in different vehicles. We then use a feature extractor to extract the features of the front-view image. We carry out experiments with ResNet18 and ResNet34, respectively\cite{he2016deep}. In order to promote the ability of the network to extract front-view features, a front-view lane detection head is added to serve as auxiliary supervision. Inspired by \cite{lin2017feature}, we design the \textit{Spatial Transformation Pyramid}, a fast multiscale spatial transformation module based on \cite{pan2020cross}. This module is responsible for the transformation from front-view features to BEV features. Finally, we forecast the lane on the plane tangent to the local road surface $P_{road}$. $P_{road}$ is the plane with $z=0$ in the road ground coordinates $\mathcal{C}_{road}=\left(x,y,z\right)$. We divide the $P_{road}$ into $s1 \times s2$ cells. Inspired by YOLO\cite{redmon2016you} and LaneNet\cite{neven2018towards}, we predict the confidence, the embedding used for clustering, the offset from the cell center to the lane in the $y$ direction of $\mathcal{C}_{road}$ and the height of each cell. In the inference, we use a fast clustering method to fuse the results of each branch to obtain 3D lanes.

\subsection{Virtual Camera}
\label{subsec:virtual}
The in/extrinsic parameters of different vehicles are various, which has a significant impact on the results of 3D lanes. Different from the methods that integrate the camera intrinsic and extrinsic parameters into the network features\cite{chen2022persformer,philion2020lift}, we realize a preprocessing method of quickly unifying the camera in/extrinsic parameters by establishing a \textit{Virtual Camera} with standard in/extrinsic parameters.

\begin{figure}
    \centering
      \includegraphics[width=.4\textwidth]{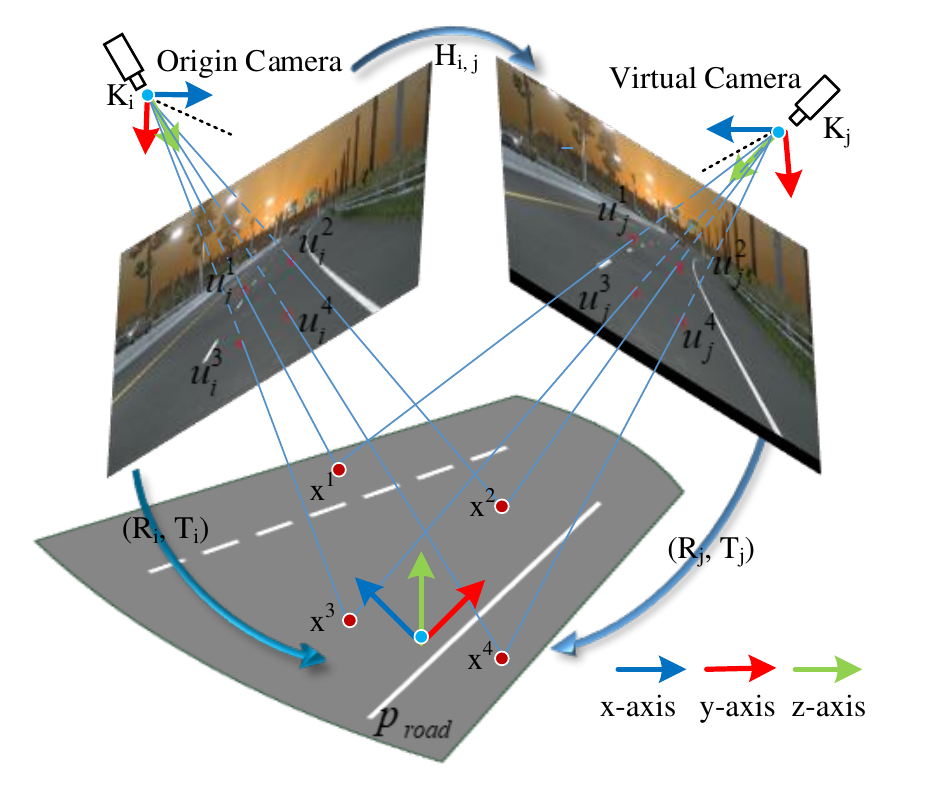}
      \caption{Schematic diagram of the \textit{Virtual Camera}. The core aspect of the \textit{Virtual Camera} is that the current camera and virtual camera are co-planar on $P_{road}$ after Inverse Perspective Mapping (IPM).}
    \label{fig:virtual}
\end{figure}

We assume $P_{road}$ to be the plane tangent to the local road surface. Because the 3D lane detection pays more attention to the plane $P_{road}$, we use the coplanarity of homography to project the image of the current camera to the view of the \textit{Virtual Camera} through the homography matrix ${{\rm{H}}_{i,j}}$. Therefore, the \textit{Virtual Camera} achieves consistency of the spatial relationship of different cameras.
As shown in Figure \ref{fig:virtual}, the intrinsic parameters ${{\rm{K}}_j}$ and extrinsic parameters ${\rm{(}}{{\rm{R}}_j},{{\rm{T}}_j})$ of the \textit{Virtual Camera} are fixed, which are derived from the mean value of the in/extrinsic parameters of the training dataset. In the training and inference stages, the homography ${{\rm{H}}_{i,j}}$ is calculated according to the camera intrinsic parameters ${{\rm{K}}_i}$ and extrinsic parameters ${\rm{(}}{{\rm{R}}_i},{{\rm{T}}_i})$ provided by the current camera and the in/extrinsic parameters of the \textit{Virtual Camera}. We refer \cite{homographies} to calculate  ${{\rm{H}}_{i,j}}$.  Firstly, we select four points ${\rm{x}}^k = {(x^k,y^k,0)^T}$ where $k=1,2,3,4$ on the BEV plane $P_{road}$. We then project them to the image of the current camera and the image of \textit{Virtual Camera} respectively to obtain ${\rm{u}}_i^k = {(u_i^k,v_i^k,1)^T}$ and ${\rm{u}}_j^k = {(u_j^k,v_j^k,1)^T}$. Finally, ${{\rm{H}}_{i,j}}$ is obtained by least square method, as shown in Eqn \ref{Enq: transform}.
\begin{equation}
{{\rm{H}}_{i,j}}{\rm{u}}_i^k = {\rm{u}}_j^k
\label{Enq: transform}
\end{equation}

During the inference, it is only necessary to perform the transformation, invoking \textit{warpPerspective} if in OpenCV, with the already obtained ${\rm{H}}_{i,j}$. 

\subsection{MLP Based Spatial Transformation Pyramid}
The depth-based \cite{philion2020lift,huang2021bevdet} and Transformer-based methods\cite{chen2022persformer,li2022bevformer} are computationally expensive and unfriendly in deployment to autopilot chips. To address the issue, we introduce a light-weight and easy-to-deploy spatial transformation module referred to \textit{View Relation Module} (VRM) \cite{pan2020cross} based on MLP. The module learns the relationships between any two pixel positions in the flattened front-view features and flattened BEV features using a view relation module $R$. However, the VRM is a fixed mapping that ignores the variations brought by different camera parameters. Fortunately, the \textit{Virtual Camera}, which unifies the in/extrinsic parameters of different cameras, makes up for this deficiency. The VRM is sensitive to the position of the front-view feature layer. We analyze the effect of different scales of front-view features in the VRM. Low resolution features are found to be more suitable for spatial transformation in the VRM through experiments. We consider that the low resolution features contain more global information. And since the MLP-based spatial transformation is a fixed mapping, the low-resolution features need fewer mapping parameters, which are easier to learn. Inspired by the FPN\cite{lin2017feature}, we design a \textit{Spatial Transformation Pyramid} based on VRM, as shown in the red box of Figure \ref{fig:backbone}. By experimental comparison, we ultimately use the 1/64 resolution feature of the input image, $S64$ and the 1/32 resolution feature named $S32$ to be transformed, respectively, and then concatenate the results of both.
\begin{equation}
\begin{split}
{f_t}[i] = concate(R_i^{S32}({f^{S32}}[1], \ldots ,{f^{S32}}[H{W^{S32}}]), \\
			  R_i^{S64}({f^{S64}}[1], \ldots ,{f^{S64}}[H{W^{S64}}]))
\end{split}
\end{equation}
where $R_i^{S32}$ denotes the VRM of $S32$, ${f_t}[i]$ denotes pixel value of BEV features, $H{W^{S32}}$ denotes the shape of $S32$, and ${f^{S32}}[j]$ denotes pixel value on $S32$.

\input{table/table_openlane_scence}

\input{table/table_openlane_new}
\input{table/table_apollo}
\input{table/table_ablation_new}
\input{table/table_head}

\input{table/table_scales}

\subsection{Key-Points Representation}
\begin{figure}
    \centering
      \includegraphics[width=7cm]{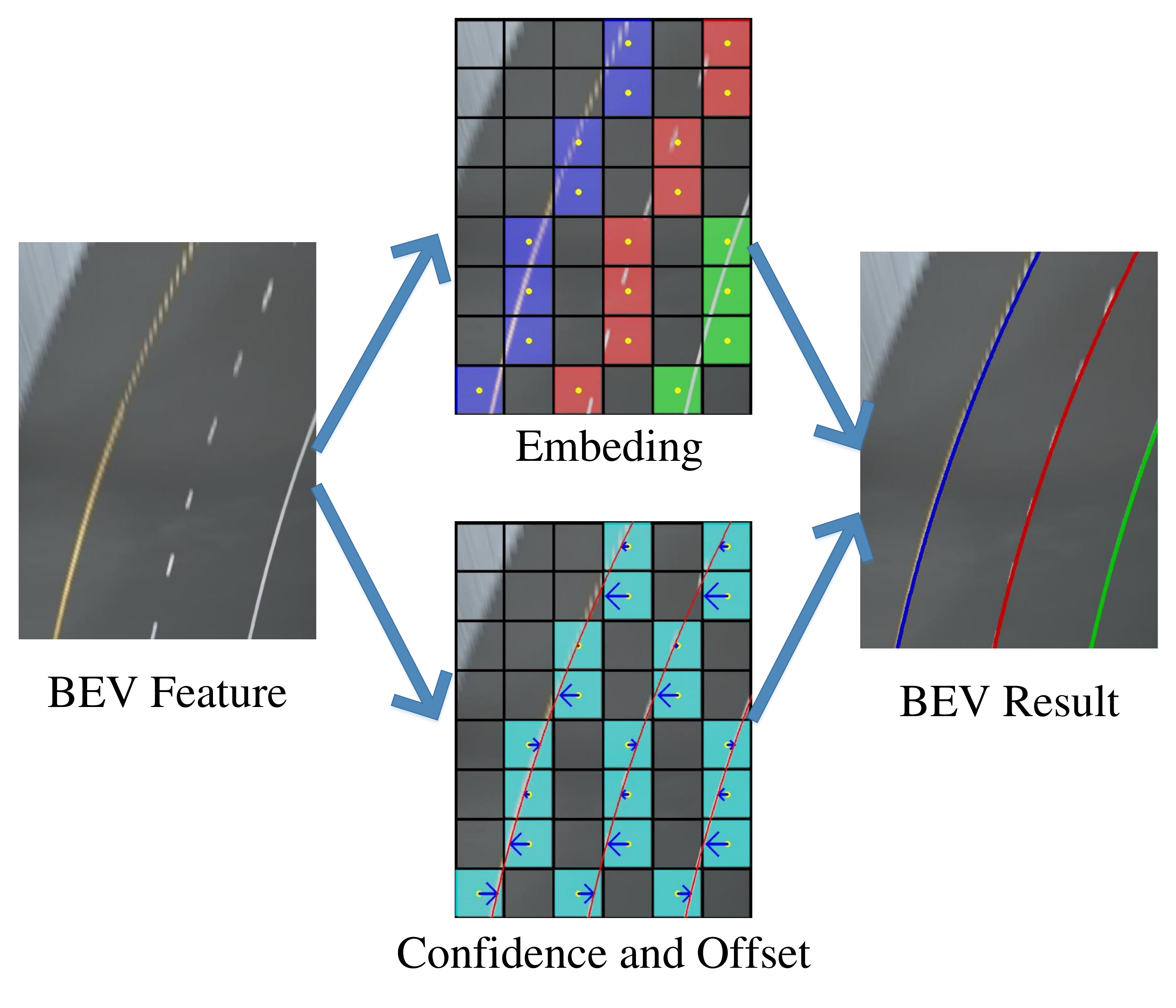}
      \caption{Schematic diagram of the \textit{Key-Points Representation}. The BEV plane $P_{road}$ is divided into $s1 \times s2$ grids. Each grid represents an area of $x \times x$ ${m^2}$. BEV features are convolved to obtain three branches, including the embedding, confidence and offset. The three branches are merged to obtain the instance-level lanes on BEV.}
    \label{fig:head}
\end{figure}

The representation of 3D lanes has a significant impact on the results of 3D lane detection. In this subsection, we propose a simple but robust representation to predict 3D lanes on BEV, referring to YOLO\cite{redmon2016you} and LaneNet\cite{neven2018towards}. As shown in Figure \ref{fig:head}, we divide the BEV plane $P_{road}$, which is the plane with $z=0$ in the road coordinates $\mathcal{C}_{road}=\left(x,y,z\right)$ into $s1 \times s2$ cells. Each cell represents $x \times x$ ($x$ defaults to 0.5$m$). We directly predict the four heads with the same resolution, including the confidence, the embedding used for clustering, the offset from the cell center to the lane in $y$ direction, and the average height of each cell.
The size of the grid cell has a great influence on 3D lane prediction. An excessively small grid cell size affects the balance of positive and negative samples in the confidence branch. However, if the cell size is too large, the embedding of different lanes will overlap. Considering the sparsity of lane tasks, we recommend that the grid cell size be $0.5 \times 0.5$ ${m^2}$ through experiments. In training and inference, we predict the lanes of (-10$m$, 10$m$) in the $y$ direction and (3$m$, 103$m$) in the $x$ direction in the road ground coordinates $\mathcal{C}_{road}=\left(x,y,z\right)$. Thus, four $200 \times 40$ resolution tensors, including confidence, embedding, offset and height are output from the 3D lane detection head. The confidence branch, embedding branch, and offset branch are merged to obtain the instance-level lanes under the BEV, as shown in Figure \ref{fig:head}.

\subsubsection{Confidence}
Similar to YOLO\cite{redmon2016you}, the confidence of lanes is a binary classification branch. Each pixel represents the confidence of the cell. If there is a lane through the cell, the confidence score of the cell is set to one. Otherwise, the confidence score is set to zero. The confidence loss can be expressed by the Binary Cross Entropy loss.
\begin{equation}
L_{conf}^{3d} = \sum\limits_i^{s1 \times s2} {({{\hat p}_i}\log {p_i} + (1 - {{\hat p}_i})\log (1 - {p_i}))}
\end{equation}
where $p_i$ denotes the probability of the confidence predicted by the model, and ${\hat p}_i$ denotes the ground truth of confidence.

\subsubsection{Offset}
Since the confidence branch does not accurately represent the location of lanes, the offset branch is responsible for predicting the precise offset from the cell center to the lane in the $y$ direction of the road ground coordinates $\mathcal{C}_{road}=\left(x,y,z\right)$. As shown in Figure \ref{fig:head}, the model predicts the offset $\Delta {y_i}$ of each cell. The offset is normalized by the Sigmoid and subtracted by 0.5 so that the range of the offset is $(-0.5, 0.5)$. The offset loss can be expressed by the MSE loss. Note that we only calculate offset for grid cells with a positive ground truth of confidence.
\begin{equation}
L_{offset}^{3d} = \sum\limits_i^{s1 \times s2} {{{\rm{1}}_{obj}}{{(\sigma (\Delta {y_i}) - \Delta {{\hat y}_i})}^2}}
\end{equation}
where ${{\rm{1}}_{obj}}$ denotes whether the lane passes through this cell. $\Delta {y_i}$ denotes offset from prediction, and $\Delta {{\hat y}_i}$ denotes offset from ground truth.

\subsubsection{Embedding}
To distinguish the lane identity of each pixel in the confidence branch, we predict the embedding of each grid cell with reference to \cite{de2017semantic, neven2018towards}. In the training stage, the distance among cell embeddings belonging to the same lane is minimized, whereas the distance among cell embeddings belonging to different lanes is maximized. In the inference of the network, we use a fast unsupervised clustering post-processing method to predict the variable number of lanes. Unlike front-view lanes that usually converge at vanishing points, 3D lanes are more suitable for the embedding clustering loss function.
\begin{equation}
L_{embed}^{3d} = L_{var}^{3d} + L_{dist}^{3d}
\end{equation}
where $L_{var}^{3d}$ denotes the loss of minimizing the mean of cell embeddings belonging to the same lane. $L_{dist}^{3d}$ denotes the loss of maximizing the variance of cell embeddings belonging to the different lanes.
\subsubsection{Lane Height}
Confidence, offset, and embedding can only predict the $x$ and $y$ of key points in the road ground coordinates $\mathcal{C}_{road}=\left(x,y,z\right)$, and thus we present a height branch that is responsible for predicting the $z$ of the key points. In the training phase of the network, we use the average height in a grid cell as ground truth. At the same time, only the grid cells with positive ground truth are counted in the loss.

\begin{equation}
L_{height}^{3d} = \sum\limits_i^{s1 \times s2} {{{\rm{1}}_{obj}}{{({h_i} - {{\hat h}_i})}^2}}
\end{equation}
where $h_i$ denotes the height of the grid cell predicted by the model, and ${\hat h}_i$ denotes the height of the grid cell from the ground truth.

\subsubsection{Total loss}
The total loss includes 3D lane losses and front-view lane losses. The front-view lane loss includes lane segmentation loss and lane embedding loss,  referred to LaneNet\cite{neven2018towards}.
\begin{equation}
\begin{split}
{L_{total}} = \lambda _{conf}^{3d}L_{conf}^{3d} + \lambda _{embed}^{3d}L_{embed}^{3d} \\
	+ \lambda _{offset}^{3d}L_{offset}^{3d} + \lambda _{height}^{3d}L_{height}^{3d}\\
     + \lambda _{seg}^{2d}L_{seg}^{2d} + \lambda _{embed}^{2d}L_{embed}^{2d}
\end{split}
\end{equation}
where $L_{seg}^{2d}$ denotes lane segmentation loss, and $L_{embed}^{2d}$ denotes lane embedding loss in the front-view.

\subsubsection{Inference}
Given the outputs from \textit{KPR}, we propose a fast unsupervised clustering method to obtain the instance-level lanes, which we refer to mean-shift from LaneNet \cite{neven2018towards}. We put the details of the algorithm in Appendix.
\begin{figure*}[t]
    \centering
      \includegraphics[width=0.9\textwidth]{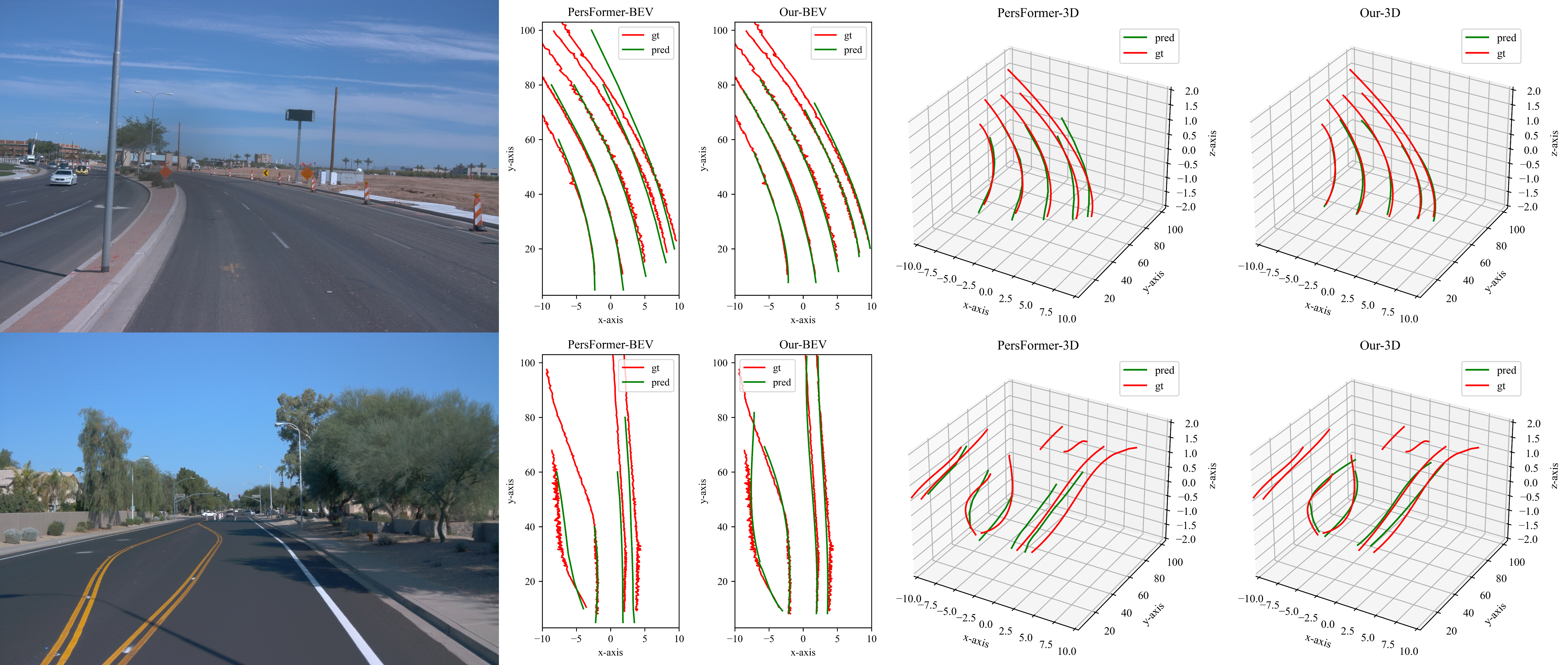}
      \caption{Qualitative results of PersFormer\cite{chen2022persformer} and BEV-LaneDet on the Openlane dataset. The first column: the input images; The second column: the results of PersFormer in BEV; The third column: the results of our method in BEV; The fourth column:  the results of PersFormer in 3D space; The fifth column: the results of our method in 3D space. The visualization results show that our method is more flexible and accurate.}
    \label{fig:show}
\end{figure*}

\section{Experiments}
\label{sec:experiment}
In order to verify the performance of our work, our model is tested on the OpenLane real-world dataset \cite{chen2022persformer} and the Apollo simulation dataset\cite{guo2020gen}. Compared with previous methods, including PersFormer\cite{chen2022persformer}, Reconstruct from Top \cite{li2022reconstruct}, Gen-LaneNet\cite{guo2020gen}, 3D-LaneNet\cite{garnett20193d}, CLGO\cite{liu2022learning}, etc., it is proven that our work can reach the state-of-the-art level in terms of F-Score and achieve competitive results in terms of X/Z error. The resolution of our input image is $576 \times 1024$.

\subsection{Evaluation Metrics and Implementation Details}

On both 3D datasets, we adopt evaluation metrics coming from Gen-LaneNet\cite{guo2020gen}, which include F-Score in various scenes and X/Z error in different areas.
\subsection{Results on OpenLane}
\label{subsec: openlane}
OpenLane contains 150,000 training frames and 40,000 test frames. In order to verify the performance of the model for every scene, the Up\&Down case, Curve case, Extreme Weather case, Intersection case, Merge\&Split case, and Night case are separated from the validation set.  Table \ref{table: openLane scence} shows the F-Score of the model in every scene. Our model trains 10 epochs in the training set and achieves the state-of-the-art performance for each scene. Table \ref{table: openLane result} shows the specific performance in terms of F-score and X/Z error for different methods. Our results are 10.6\% higher than the state-of-the-art work\cite{chen2022persformer} in terms of F-Score. The detailed visualization can be found in Appendix. Moreover, since the 3D ground truth of OpenLane is synthesized by LiDAR, our work does not show much advantage in X error. However, our work shows a great advantage with respect to X error for the Apollo dataset\cite{guo2020gen}.

\subsection{Results on Apollo 3D synthetic}
\label{subsec: apollo}
The Apollo dataset \cite{guo2020gen} includes 10,500 discrete frames of monocular RGB images and their corresponding 3D lanes ground truth, which is split into three scenes: balanced, rarely observed, and visual variation scenes. Each scene contains independent training sets and test sets. It is noted that Apollo does not provide specific extrinsic parameters of the camera. We calculate the extrinsic parameters of the camera through the height and pitch of the camera provided by the dataset. In Table \ref{table: apollo}, we provide a comparison between the previous works and our work. Our model has trained 80 epochs on these datasets. The F-Score and X error of our work both reach the state-of-the-art with respect to Apollo. However, since our work is more focused on the BEV plane, our work does not perform well in terms of Z error. We will improve this shortcoming in the future.

\subsection{Ablation Study}
\label{sec: ablation}
The experiments in this section will be carried out on the OpenLane, and the evaluation metrics are still based on Gen-LaneNet\cite{guo2020gen}. Our baseline uses ResNet34 as the backbone, no \textit{Virtual Camera}(VC), VRM \cite{pan2020cross} as the spatial transformation module, \textit{Key-Points Representation} (KPR) with offset and a grid size of 0.2 for 3D lanes, and no 2D auxiliary supervision. We prove the effectiveness of our methods by adding three modules: VC, \textit{Spatial Transformation Pyramid} (STP), and 2D auxiliary supervision. as shown in Table \ref{table: ablation}. Also, the table demonstrates the effect of changing the cell size from 0.2 to 0.5 in KPR, which we will discuss in detail in the following experiments. Meanwhile, we demonstrate the speed of different backbone running on the Tesla-V100.

At the same time, in order to verify the effect of the offset and grid cell sizes on KPR, we add Table \ref{table: head}. In the experiments without offset, the F-Score keeps growing as the cell size grows. However, as the cell size increases, the X error also increases. When offset is added, the X error returns to the normal level for the large cell. Meanwhile, the larger the cell size is, the smaller FLOPs is. The experimental results demonstrate that the model achieves the best metrics when gird cell size is $0.5 \times 0.5$ ${m^2}$ with offset.
Moreover, we explore the influence of the position of the front-view feature layer on the \textit{View Relation Module}\cite{pan2020cross} in Table \ref{table: scales}. Experiments show that the \textit{Spatial Transformation Pyramid} achieves the best results when fusing the 64x down-sampling features and 32x down-sampling features of the original image.

\section{Conclusions}
In this paper, we propose BEV-LaneDet, a simple but effective 3D lane detection method. We present a \textit{Virtual Camera} to guarantee the consistency of the spatial relationship of front-facing cameras in different vehicles, and we have proven its effectiveness through experiments. Moreover, we demonstrate experimentally that the \textit{Spatial Transformation Pyramid}, which is a robust and chip-friendly module for spatial transformation, is effective. As shown in experiments, \textit{Key-Points Representation} is a simple but effective module, which is more suitable to represent the diversity of lane structures. At last, we believe that our method can facilitate additional on-road 3D tasks.

{
	\small
	\bibliographystyle{ieee_fullname}
	\bibliography{egbib}
}
\clearpage
\appendix
\section{Appendix}

\subsection{Inference}
In the inference phase, the four branches of the network output, including the confidence $M_{conf}$, the embedding $M_{emb}$, the offset $M_{off}$, and the lane height $M_Z$, are synthesized into the instance-level lanes by a fast unsupervised clustering method which we refer to mean-shift. As shown in Alg \ref{alg:Inference post-processing}. This process consists of four steps: 1) Filtering the positive points of confidence mask $M_{conf}$ by the $S_{threshold}$ to obtain the $E_{list}$. 2) Clustering points of $M_{emb}$ to obtain the clustering points of each lane $R_{point}$  and cluster centers of each lane $R_{center}$ by inter-class distance $D_{gap}$. 3) Adding offset $M_{off}$ and lane height $M_Z$  to points in each lane to obtain 3D lanes $R_{lines}$. 4) Fitting the key points of the lanes $R_{lines}$ to yield the lane equations $R_{fit}$.

\input{table/algorithm}

\subsection{Visualization}
We analyse the comparison between BEV-LaneDet and PersFormer\cite{chen2022persformer} in different scenarios by visualization. In Figure \ref{fig:show}, the first column is the input images; the second column is the results of PersFormer on BEV; the third column is the results of our method on BEV; the fourth column is the results of PersFormer in 3D space; the fifth column is the results of our method in 3D space. The visualization results demonstrate that our method is more stable and accurate in different scenarios. At the same time, our method is more suitable to represent the diversity of lane structures compared with PersFormer.

\begin{figure*}[t]
    \centering
      \includegraphics[width=0.9\textwidth]{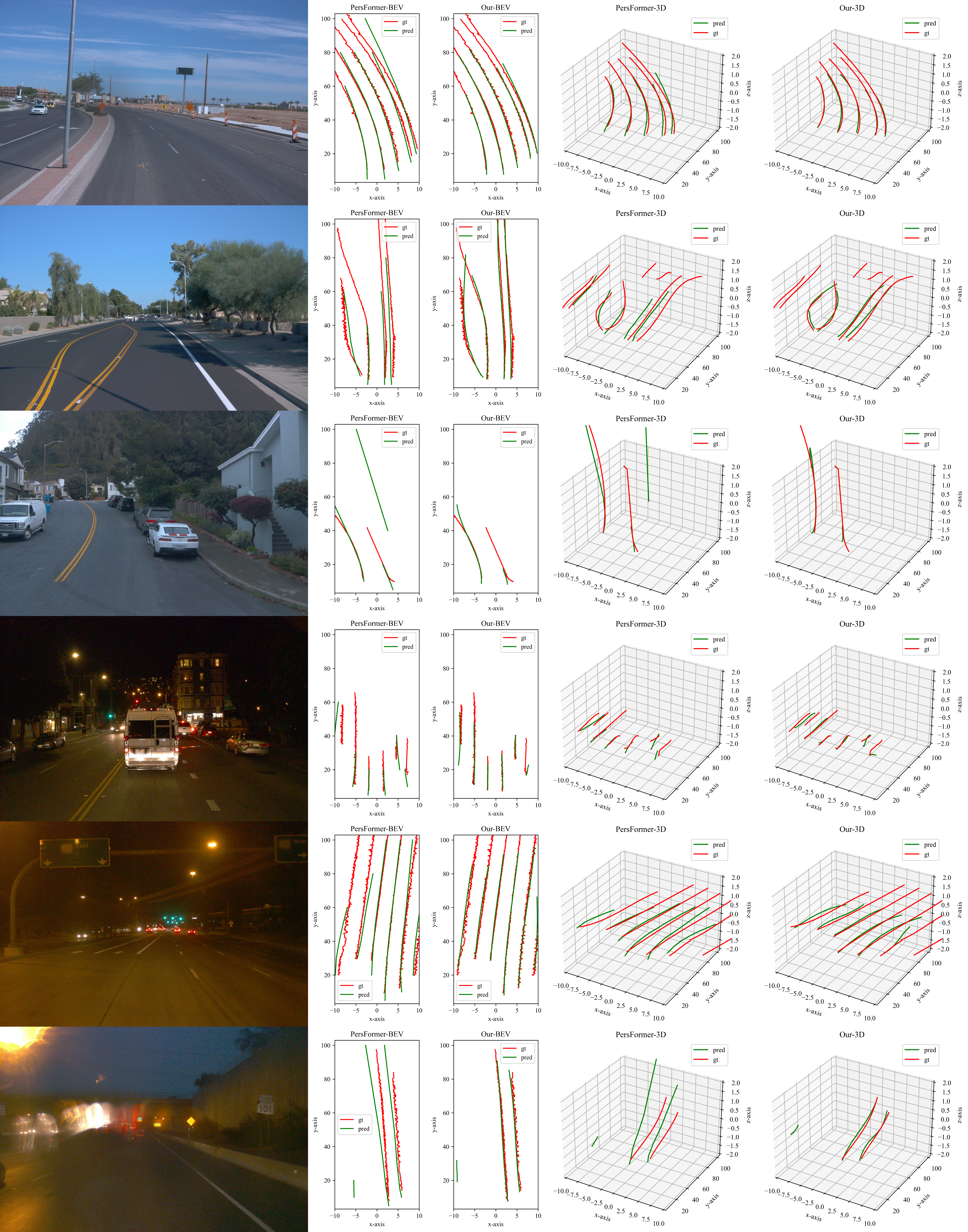}
      \caption{Qualitative results of PersFormer\cite{chen2022persformer} and BEV-LaneDet in different scenarios of the OpenLane dataset. First row: Curve; second row: Merge\&Split; third row: Up\&Down; fourth row: Night; fifth row: Intersection; sixth row: Backlight.}
    \label{fig:show}
\end{figure*}

\end{document}

%% file: chapter/introduction.tex
%
 As one of the fundamental guarantees for autonomous driving, lane detection has recently received much attention from researchers. Robust lane detection in real-time is one of the foundations for advanced autonomous driving, which can provide substantial amounts of useful information for Autonomous Driving Systems (ADS), vehicle self-control, localization, and map construction.
\begin{figure}[ht]
    \centering
      \includegraphics[width=.5\textwidth]{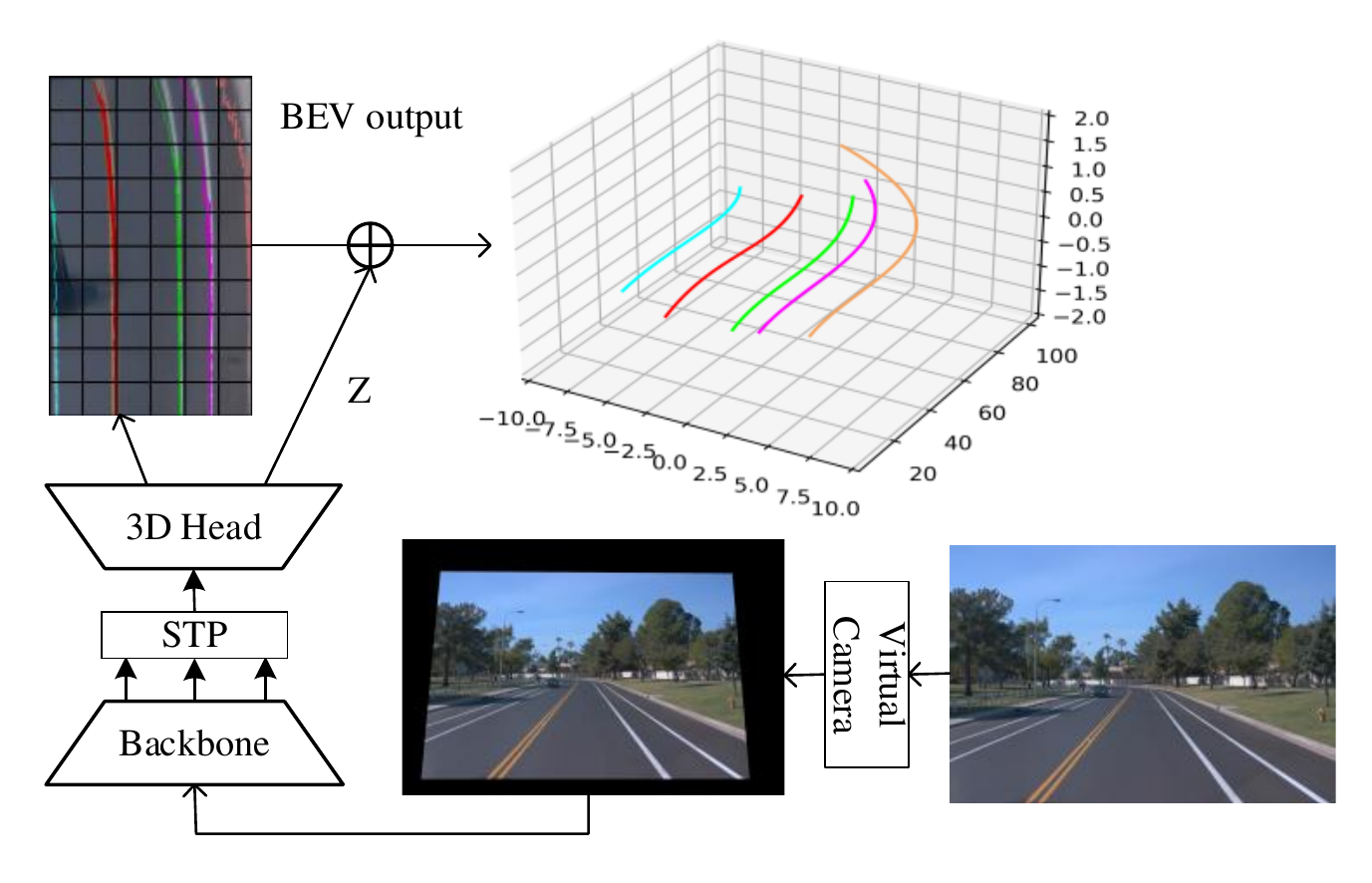}
      \caption{\textbf{End-to-end framework illustrated.} The original image in bottom-right is transformed to an input image by the \textit{Virtual Camera} module. The input image is then encoded into front-view features by the backbone. Multiscale features from the backbone are put into the \textit{Spatial Transformation Pyramid} (STP) to obtain BEV features. Given the BEV features, the 3D Head called \textit{Key-Points Representation} generates BEV output and Z, the height of BEV lanes. At last, with the BEV output and Z, we can obtain 3D lanes.}
    \label{fig:struct}
\end{figure}

2D lane detection methods have demonstrated remarkable performances \cite{cordts2016cityscapes,pan2018spatial,neven2018towards,liu2021condlanenet}. Moreover, their outputs are usually projected to the flat ground plane by Inverse Perspective Transformation (IPM) with the camera in/extrinsic parameters, and then curve fitting is performed to obtain the BEV lanes. However, the pipeline might cause other problems in the actual driving process \cite{bai2018deep,neven2018towards} for challenging situations like uphill and downhill.

In order to overcome these problems, more recent methods \cite{garnett20193d,efrat20203d,guo2020gen,li2022reconstruct,chen2022persformer} have started to focus on the more complicated 3D lane perception domain. There are two significant challenges in 3D lane detection: an efficient spatial transformation module to obtain BEV features and a robust representation for 3D lane structures. The acquisition of BEV features is heavily dependent on camera in/extrinsic parameters, and previous methods have chosen to incorporate camera in/extrinsic parameters into the network to obtain BEV features. Moreover, unlike obstacles on the road, lane structures are slender and diverse. These methods \cite{garnett20193d, guo2020gen, chen2022persformer} carefully design the 3D anchor representation for lane structures with strong priors. However, they lack sufficient flexibility in some specific scenarios, as shown in Figure \ref{fig:show}. Moreover, the anchor-free method \cite{efrat20203d} proposes a 3D lane representation based on the hypothesis that a line segment in each predefined tile is straight. This representation is complicated and inaccurate.

Towards the issues, we introduce BEV-LaneDet, an efficient and real-time pipeline that achieves 3D lane detection from a single image, as shown in Figure \ref{fig:struct}. Different from the methods of incorporating camera in/extrinsic parameters into the network to get BEV features, we establish a \textit{Virtual Camera}, which is applied to images directly. The module unifies the in/extrinsic parameters of front-facing cameras in different vehicles by the homography method\cite{homographies} based on BEV. This module guarantees the consistency of the spatial relationship of front-facing cameras in different vehicles and reduces variance in data distribution. Therefore, it can effectively promote the learning procedure due to the unified visual space. We also propose a \textit{Key-Points Representation} as our 3D lane representation. We demonstrate that it is a simple but effective module to represent 3D lanes and is more expandable for complicated lane structures in some special scenarios. Moreover, the cost of computation and the chip's friendliness are also crucial factors in autonomous driving. Therefore, a light-weight and easy-to-deploy spatial transformation module based on MLP is our preference. Meanwhile, inspired by FPN\cite{lin2017feature}, we present the \textit{Spatial Transformation Pyramid}, which transforms multiscale front-view features to BEV and provides robust BEV features for 3D lane detection. In our experiments, we perform extensive studies to confirm that our BEV-LaneDet significantly outperforms the state-of-the-art PersFormer \cite{chen2022persformer} in terms of F-Score, being 10.6\% higher on the OpenLane real-world test set \cite{chen2022persformer}  and 4.0\% higher on the Apollo simulation test set \cite{guo2020gen} with a speed of 185 FPS.

\textbf{In summary, our main contributions are three-fold:} \textbf{1)} \textit{Virtual Camera}, a novel preprocessing module to unify the in/extrinsic parameters of cameras, ensuring data distribution consistency. \textbf{2)} \textit{Key-Points Representation}, a simple but effective representation of 3D lane structures. \textbf{3)} \textit{Spatial Transformation Pyramid}, a light-weight and easy-to-deploy architecture based on MLP to realize transformation from multiscale front-view features to BEV. Experiments demonstrate that our BEV-LaneDet achieves the state-of-the-art performance compared to other 3D lane detection algorithms.

%% file: chapter/related_work.tex
\textbf{2D Lane Detection.}
In recent years, there have been significant advancements in the field of 2D lane detection using deep neural networks (DNN). These works are divided into four categories according to pixel-wise segmentation, row-wise methods, anchor-based methods, and curve parameters. Some recent works \cite{pan2018spatial,zheng2021resa,li2021hdmapnet,neven2018towards} consider 2D lane detection as a segmentation task based on pixel-wise, which the computing cost is expensive. 
Some methods \cite{qin2020ultra,liu2021condlanenet,yoo2020end} focus on the row-wise level to detect the 2D lanes. By setting the row anchors in the row direction and setting the grid cells in the column direction to model the 2D lanes on the image space, row-wise methods greatly improve the speed of inference. \cite{tabelini2021keep, su2021structure} represent the lane structures with predefined anchors and regress offsets between sampled points and predefined anchor points to predict 2D lanes. These methods lack sufficient flexibility to accommodate complex lanes due to fixed anchors design.
\cite{feng2022rethinking,tabelini2021polylanenet} argue that the lane can be fitted by specific curve parameters on the 2D image space. So it is proposed that the 2D lane detection can be converted into the problem of curve parameter regression by detecting the starting point, ending point, and curve parameters. However, these methods need to combine camera intrinsic and extrinsic parameters for IPM projection to the ground in post-processing, which is based on the flat ground hypothesis. As mentioned in Section \ref{sec:introduction}, this pipeline is not suitable for complicated road scenarios.

\textbf{3D Lane Detection.}
In order to obtain more accurate road cognition results, many researchers have turned their attention to lane detection in 3D space. \cite{garnett20193d,guo2020gen,chen2022persformer,li2022reconstruct} present remarkable results to prove the feasibility of using a CNN network for 3D lane detection in monocular images. 3D-LaneNet \cite{garnett20193d} firstly introduces a unified network for encoding 2D image information, spatial transformation and 3D lane detection in two path-ways: the image-view path encodes features from the 2D image, while the top-view path provides translation-invariant features for 3D lane detection. 3D-LaneNet+ \cite{efrat20203d} constructs the shape of lane segments in the predefined grid cells based on the straight segment hypothesis, which is complicated and might cause an error between the predicted segments and the actual lanes. Gen-LaneNet \cite{guo2020gen} proposes an extensible two-stage framework that separates the image segmentation subnetwork and the geometry encoding subnetwork. PersFormer \cite{chen2022persformer} proposes a unified 2D and 3D lane detection framework and introduces Transformer \cite{vaswani2017attention} into the spatial transformation module to obtain more robust features. It also proposes a real-scene-based and large-scale annotated 3D lane dataset, OpenLane. These methods deploy intra-network feature maps IPM projection with camera in/extrinsic parameters, implicitly or explicitly. Different from applying IPM projection to features, we construct a \textit{Virtual Camera} module, trying to project all images onto the view of a standard virtual common camera. This approach ensures that the distribution of images is as consistent as possible. The distribution here includes the position, angle, and height of the camera from the ground. As for 3D lane representation, they carefully design 3D anchors with strong priors, which are complicated and lack expressiveness for some specific scenarios, as shown in Figure \ref{fig:show}. We present our 3D lane representation, which is simple but robust, and it is more expandable for special scenarios.

\textbf{Spatial Transformation.}
A vital module of 3D lane detection is the spatial transformation from front-view features to BEV features. The spatial transformation module\cite{jaderberg2015spatial} is a trainable module that is flexibly inserted into the CNN to implement the spatial transformation of the input features, and it is suitable for converting front-view features into BEV geometric features. There are four kinds of commonly used spatial transformation modules. IPM-based methods \cite{reiher2020sim2real,garnett20193d,guo2020gen} rely heavily on the camera in/extrinsic parameters and ignore ground surface undulations and vehicle vibrations. The MLP-based methods \cite{pan2020cross, li2021hdmapnet} are fixed spatial mapping, which are difficult to be integrated with the camera in/extrinsic parameters, resulting in poor performance. However, it is chip-friendly and rapid. Transformer-based spatial transformation modules\cite{chen2022persformer,li2022bevformer} are more robust, but they are not easy to deploy into autopilot chips due to the large amount of computation. The spatial transformation methods based on depth\cite{philion2020lift,huang2021bevdet} have a large amount of calculation and are thus not suitable for deployment. In combination with the \textit{Virtual Camera} module, we can overcome the drawbacks of the MLP-based approaches and apply them to our method. We also introduce a feature pyramid inspired by FPN\cite{lin2017feature} to provide more robust BEV features.


%% file: table/table_openlane_scence.tex
\begin{table*}[t]
\centering

\caption{Comparison with other open-sourced 3D methods on the OpenLane. Our method achieves the best F-Score on the entire validation set and every scenario set.~\\}
\label{table: openLane scence}
\scalebox{1.0}{
\begin{tabular}{llllllll}
\hline
\textbf{Method} &
  \textbf{All} &
  \textbf{\begin{tabular}[c]{@{}l@{}}Up\&\\ Down\end{tabular}} &
  \textbf{Curve} &
  \textbf{\begin{tabular}[c]{@{}l@{}}Extreme\\ Weather\end{tabular}} &
  \textbf{Night} &
  \textbf{Intersection} &
  \textbf{\begin{tabular}[c]{@{}l@{}}Merge\&\\ Split\end{tabular}} \\ \hline
3D-LaneNet\cite{garnett20193d}  & 40.2        & 37.7          & 43.2          & 43            & 39.3          & 29.3          & 36.5          \\
Gen-LaneNet\cite{guo2020gen} & 29.7        & 24.2          & 31.1          & 26.4          & 19.7          & 19.7          & 27.4          \\
PersFormer\cite{chen2022persformer}   & 47.8        & 42.4          & 52.8          & 48.7          & 37.9          & 37.9          & 44.6          \\
\rowcolor[HTML]{C0C0C0} 
Ours        & \textbf{58.4} & \textbf{48.7} & \textbf{63.1} & \textbf{53.4} & \textbf{53.4} & \textbf{50.3} & \textbf{53.7} \\ \hline
\end{tabular}
}
\end{table*}

%% file: table/table_openlane_new.tex
\begin{table*}[t]
\centering
\caption{Comprehensive 3D lane evaluation under different metrics. Our Method outperforms previous 3D methods on the metrics of F-Score and speed.~\\}

\label{table: openLane result}
\scalebox{1.0}{
\begin{tabular}{llllllll}
\hline
\textbf{Method} & \textbf{F-Score} & \textbf{X error near} & \textbf{X error far} & \textbf{Z error near} & \textbf{Z error far} & \textbf{Pytorch} & \textbf{TensorRT}\\ \hline
3D-LaneNet\cite{garnett20193d}  & 40.2        & \textbf{0.278} & 0.823          & \textbf{0.159} & 0.714   & -     & -      \\
Gen-LaneNet\cite{guo2020gen} & 29.7        & 0.309          & 0.877          & 0.16           & 0.75     & 54FPS     & -          \\
PersFormer\cite{chen2022persformer}  & 47.8        & 0.322          & 0.778          & 0.213          & 0.681   & 21FPS     & -       \\
\rowcolor[HTML]{C0C0C0} 
Ours        & \textbf{58.4} & 0.309          & \textbf{0.659} & 0.244          & \textbf{0.631} & \textbf{102FPS}  & \textbf{185FPS}\\ \hline
\end{tabular}
}

\end{table*}

%% file: table/table_apollo.tex
\begin{table*}[t]
\centering
\caption{There are results of different models on Apollo 3D Lane Synthetic. Compared with 3D-LaneNet\cite{garnett20193d}, Gen-LaneNet\cite{guo2020gen}, 3D-LaneNet(1/att)\cite{jin2021robust}, Gen-LaneNet(1/att)\cite{jin2021robust}, CLGO\cite{liu2022learning}, Reconstruct from Top\cite{li2022reconstruct}, and PersFormer\cite{chen2022persformer}, our model is the best on F-Score and X error.}
\label{table: apollo}
\begin{tabular}{lllllllll}
 &
  \multicolumn{6}{l}{\textbf{}} \\ \hline
\textbf{Scene} &
  \textbf{Method} &
  \textbf{F-Score} &
  \begin{tabular}[c]{@{}l@{}}\textbf{X error near}\end{tabular} &
  \begin{tabular}[c]{@{}l@{}}\textbf{X error far}\end{tabular} &
  \begin{tabular}[c]{@{}l@{}}\textbf{Z error near}\end{tabular} &
  \begin{tabular}[c]{@{}l@{}}\textbf{Z error far}\end{tabular} \\ \hline
 &
  3D-LaneNet\cite{garnett20193d} &
  86.4 &
  0.068 &
  0.477 &
  0.015 &
  \textbf{0.202} \\
 &
  Gen-LaneNet\cite{guo2020gen} &
  88.1 &
  0.061 &
  0.496 &
  0.012 &
  0.214 \\
 &
  3D-LaneNet(1/att)\cite{jin2021robust} &
  91 &
  0.082 &
  0.439 &
  0.011 &
  0.242 \\
 &
  Gen-LaneNet(1/att)\cite{jin2021robust} &
  90.3 &
  0.08 &
  0.473 &
  0.011 &
  0.247 \\
 &
  CLGO\cite{liu2022learning} &
  91.9 &
  0.061 &
  0.361 &
  0.029 &
  0.25 \\
 &
  Reconstruct from Top\cite{li2022reconstruct}&
  91.9 &
  0.049 &
  0.387 &
  \textbf{0.008} &
  0.213 \\
 &
  PersFormer\cite{chen2022persformer} &
  92.9 &
  0.054 &
  0.356 &
  0.01 &
  0.234 \\
\multirow{-8}{*}{\begin{tabular}[c]{@{}l@{}}Balanced\\ Scence\end{tabular}} &
  \cellcolor[HTML]{C0C0C0}Ours &
  \cellcolor[HTML]{C0C0C0}\textbf{98.7} &
  \cellcolor[HTML]{C0C0C0}\textbf{0.016} &
  \cellcolor[HTML]{C0C0C0}\textbf{0.242} &
  \cellcolor[HTML]{C0C0C0}0.02 &
  \cellcolor[HTML]{C0C0C0}0.216 \\ \hline
 &
  3D-LaneNet\cite{garnett20193d} &
  72 &
  0.166 &
  0.855 &
  0.039 &
  \textbf{0.521} \\
 &
  Gen-LaneNet\cite{guo2020gen} &
  78 &
  0.139 &
  0.903 &
  0.03 &
  0.539 \\
 &
  3D-LaneNet(1/att)\cite{jin2021robust} &
  84.1 &
  0.289 &
  0.925 &
  0.025 &
  0.625 \\
 &
  Gen-LaneNet(1/att)\cite{jin2021robust} &
  81.7 &
  0.283 &
  0.915 &
  0.028 &
  0.653 \\
 &
  CLGo\cite{liu2022learning} &
  86.1 &
  0.147 &
  0.735 &
  0.071 &
  0.609 \\
 &
  Reconstruct from Top\cite{li2022reconstruct} &
  83.7 &
  0.126 &
  0.903 &
  \textbf{0.023} &
  0.625 \\
 &
  PersFormer\cite{chen2022persformer} &
  87.5 &
  0.107 &
  0.782 &
  0.024 &
  0.602 \\
\multirow{-8}{*}{\begin{tabular}[c]{@{}l@{}}Rarely\\ Observed\end{tabular}} &
  \cellcolor[HTML]{C0C0C0}Ours &
  \cellcolor[HTML]{C0C0C0}\textbf{99.1} &
  \cellcolor[HTML]{C0C0C0}\textbf{0.031} &
  \cellcolor[HTML]{C0C0C0}\textbf{0.594} &
  \cellcolor[HTML]{C0C0C0}0.04 &
  \cellcolor[HTML]{C0C0C0}0.556 \\ \hline
 &
  3D-LaneNet\cite{garnett20193d} &
  72.5 &
  0.115 &
  0.601 &
  0.032 &
  \textbf{0.23} \\
 &
  Gen-LaneNet\cite{guo2020gen} &
  85.3 &
  0.074 &
  0.538 &
  0.015 &
  0.232 \\
 &
  3D-laneNet(1/att)\cite{jin2021robust} &
  85.4 &
  0.118 &
  0.559 &
  0.018 &
  0.29 \\
 &
  \cellcolor[HTML]{FFFFFF}{\color[HTML]{000000} Gen-LaneNet(1/att)\cite{jin2021robust}} &
  \cellcolor[HTML]{FFFFFF}{\color[HTML]{000000} 86.8} &
  \cellcolor[HTML]{FFFFFF}{\color[HTML]{000000} 0.104} &
  \cellcolor[HTML]{FFFFFF}{\color[HTML]{000000} 0.544} &
  \cellcolor[HTML]{FFFFFF}{\color[HTML]{000000} 0.016} &
  \cellcolor[HTML]{FFFFFF}{\color[HTML]{000000} 0.294} \\
 &
  CLGo\cite{liu2022learning} &
  87.3 &
  0.084 &
  0.464 &
  0.045 &
  0.312 \\
 &
  Reconstruct from Top\cite{li2022reconstruct} &
  89.9 &
  0.06 &
  0.446 &
  \textbf{0.011} &
  0.235 \\
 &
  PersFormer\cite{chen2022persformer} &
  89.6 &
  0.074 &
  0.43 &
  0.015 &
  0.266 \\
\multirow{-8}{*}{\begin{tabular}[c]{@{}l@{}}Vivual \\ Variants\end{tabular}} &
  \cellcolor[HTML]{C0C0C0}Ours &
  \cellcolor[HTML]{C0C0C0}\textbf{96.9} &
  \cellcolor[HTML]{C0C0C0}\textbf{0.027} &
  \cellcolor[HTML]{C0C0C0}\textbf{0.32} &
  \cellcolor[HTML]{C0C0C0}0.031 &
  \cellcolor[HTML]{C0C0C0}0.256 \\ \hline
\end{tabular}
\end{table*}

%% file: table/table_ablation_new.tex

\begin{table}[]
\caption{Ablation studies on OpenLane, all with 2D supervision. VC: \textit{Virtual Camera}; STP: \textit{Spatial Transform Pyramid}; KPR: \textit{Key-Points Representation}; R34: ResNet34; R18: ResNet18. }
\centering
\label{table: ablation}
\scalebox{0.85}{
\begin{tabular}{l|lll|lll}
\hline
\textbf{Backbone} & \textbf{VC} & \textbf{STP} & \textbf{KPR} & \textbf{F-Score} & \textbf{X error} & \textbf{FPS}\\ \hline
R34   &            &              &              & 51.2             & 0.37/0.79       & -          \\
R34   &\checkmark           &              &              & 54.5(+3.3)       & 0.32/0.69      & -           \\
R34   &            & \checkmark            &              & 53.2(+2.0)                 & 0.37/0.79       & -          \\
R34   &            &              & \checkmark            & 53.5(+2.3)                 & 0.37/0.76       & -          \\
R34   &            & \checkmark            & \checkmark            & 55.3(+4.1)                 & 0.36/0.79         & -        \\
R34   &\checkmark           &              & \checkmark            & 56.7(+5.5)                 & 0.31/0.69         & -        \\
R34   &\checkmark           & \checkmark            & \checkmark            & 58.4(+7.2)                 & 0.31/0.66        & 185         \\
R18   &\checkmark           & \checkmark            & \checkmark            & 57.8(+7.2)                 & 0.32/0.70        & 272         \\ \hline
\end{tabular}
}
\end{table}

%% file: table/table_head.tex

\begin{table}[]
\centering
\caption{Impact of cell size and offset.}
\label{table: head}
\begin{tabular}{@{}lllll@{}}
\toprule
\textbf{\begin{tabular}[c]{@{}l@{}}Cell size \\ and offset\end{tabular}} & \textbf{F-Score} & \textbf{\begin{tabular}[c]{@{}l@{}}X error \\ near\end{tabular}} & \textbf{\begin{tabular}[c]{@{}l@{}}X error \\ far\end{tabular}} & \textbf{GFLOPs} \\ \midrule
0.05m                                                                    & 43.2             & 0.3415                                                           & 0.770                                                           & 735.31          \\
0.2m                                                                     & 55.7             & 0.321                                                            & 0.701                                                           & 89.42           \\
0.5m                                                                     & 57.9             & 0.429                                                            & 0.734                                                           & 53.05           \\
0.5m + offset                                                            & \textbf{58.4}    & \textbf{0.309}                                                   & \textbf{0.659}                                                  & 53.25           \\
1m                                                                       & 56.8             & 0.607                                                            & 0.856                                                           & 47.97           \\
1m + offset                                                              & 57.7             & 0.317                                                            & 0.671                                                           & 48.08           \\ \bottomrule
\end{tabular}
\end{table}

%% file: table/table_scales.tex
\begin{table}[]
\caption{The comparison of the different scales in front-view feature layers during spatial transformation. $S32$ represents 32x downsampling of the input image. $S32+S64$ represents the concatenation of 64x downsampling and 32x downsampling of the input image.}

\centering
\label{table: scales}
\begin{tabular}{llll}
\hline
\textbf{\begin{tabular}[c]{@{}l@{}}Combination \\ of scales\end{tabular}} & \textbf{F-Score} & \textbf{\begin{tabular}[c]{@{}l@{}}X error \\ near\end{tabular}} & \textbf{\begin{tabular}[c]{@{}l@{}}X error \\ far\end{tabular}} \\ \hline
S8                                                                        & 53.2             & 0.341                                                            & 0.682                                                           \\
S16                                                                       & 57.1             & 0.340                                                            & 0.696                                                           \\
S32                                                                       & 57.6             & 0.323                                                            & 0.682                                                           \\
S64                                                                       & 56.7             & 0.322                                                            & 0.697                                                           \\
S128                                                                      & 54.5             & 0.325                                                            & 0.676                                                           \\
S32+S64                                                                   & \textbf{58.4}    & \textbf{0.309}                                                   & 0.659                                                           \\
S32+S64+S128                                                              & 58.3             & 0.315                                                            & \textbf{0.656}                                                   \\ \hline
\end{tabular}
\end{table}

%% file: table/algorithm.tex
\renewcommand{\algorithmicrequire}{\textbf{Input:}}
\renewcommand{\algorithmicensure}{\textbf{Output:}}
\begin{algorithm}\footnotesize
\caption{Post-processing algorithm of our method}
\label{alg:Inference post-processing}
\begin{algorithmic}[1]
\REQUIRE $M_{conf}, M_{emb}, M_{off}, M_{Z}\leftarrow Model(I_v)$;
\ENSURE 3D lanes after fitting, $R_{fit}$.
\STATE Filtering the point of confidence mask by $S_{threshold}$ to get the $E_{list}$:

\FOR{$x=0$; $x<M_{conf}.cols$; $x++$}
\FOR{$y=0$; $y<M_{conf}.rows$; $y++$}
\IF{$M_{conf}[x, y] >= S_{threshold}$}
\STATE $E_{list}.append([x, y, M_{emb}[:, y, x]])$;
\ENDIF
\ENDFOR
\ENDFOR
\STATE {Clustering points to get the $R_{point}$  and $R_{center}$ by $D_{gap}$}:
\FOR{$i=0$; $i<E_{list}.length$; $i++$}
\STATE $x, y, value = E_{list}[i]$;
\STATE $min\_gap = D_{gap} + 1$;
\STATE $min\_cid = -1$;
\FOR{$j=0$; $j<R_{center}.length$; $j++$}
\STATE $center\_id, (center, num) = R_{center}[j]$;
\STATE $diff = Euclidean(value, center)$;
\IF{$diff < min\_gap$}
\STATE $min\_gap = diff$;
\STATE $min\_cid = center\_id$;
\ENDIF
\ENDFOR
\IF{$min\_gap < D_{gap}$}
\STATE $R_{point}.append([x, y, min\_cid])$;
\STATE $center, num = R_{center}[min\_cid]$;
\STATE $R_{center}[min\_cid] = [(center \times num + value) / (num + 1), num + 1]$;
\ELSE 
\STATE $R_{center}.append([value, 1])$;
\STATE $R_{point}.append([x, y, R_{center}.length - 1])$
\ENDIF
\ENDFOR
\STATE Adding offset to points in each lane to get $R_{lanes}$:
\FOR{$k = 0$; $k < R_{point}.length$; $k++$}
\STATE $x, y, id = R_{point}[i]$;
\STATE $off\_y = M_{off}[:, y, x]$;
\STATE $z = M_{Z}[y, x]$;
\STATE $R_{lanes}[id].append([x, y+off\_y, z])$;
\ENDFOR
\STATE $R_{fit} = FitFunc(R_{lanes})$;

\end{algorithmic}
\end{algorithm}